\documentclass[]{article}
\usepackage{named}
\usepackage{graphicx} %
\usepackage{amsmath} %
\usepackage{amsthm} %
\makeatletter
\def\url@leostyle{%
  \@ifundefined{selectfont}{\def\UrlFont{\sf}}{\def\UrlFont{\small\ttfamily}}}
\makeatother
\usepackage{hyperref}  

\title{Informal Physical Reasoning Processes%
}

\author{Kurt Ammon\footnote{Correspondence to kurtammon[at]csyst.org. Comments are welcome.}} 
\date{} %

\newtheorem{rl}{Rule}
\newcommand{\br}{\begin{rl}\rm}
\newcommand{\er}{\end{rl}}
\newcommand{\all}{\mbox{\it all\/}}
\newcommand{\cf}{\mbox{\it closed-formula\/}}
\newcommand{\consistent}{\mbox{\it consistent\/}}
\newcommand{\omegaconsistent}{\mbox{\it $\omega$-consistent\/}}

\newcommand{\provable}{\mbox{\it provable\/}}
\newcommand{\nt}{\mbox{\it not\/}}
\newcommand{\ex}{\mbox{\it there-is\/}}
\newcommand{\bt}{\vspace{-.1cm}\begin{tabular}}
\newcommand{\et}{\end{tabular}\vspace{-.1cm}}

\newcommand{\fnt}{\mbox{\it fnt\/}}

\newcommand{\App}{${\rm A}_{p}(\boldsymbol{p})$}             
\newcommand{\Apstrokepstroke}{${\rm A}_{p'}(\boldsymbol{p'})$}             

\newcommand{\bc}{\begin{center}}            
\newcommand{\ec}{\end{center}}
\newcommand{\bqu}{\begin{quote}}
\newcommand{\equ}{\end{quote}}
\newcommand{\bci}{\begin{center} \begin{minipage}{7.5cm} }
\newcommand{\eci}{\end{minipage} \end{center} }
\newcommand{\be}{\begin{enumerate}}
\newcommand{\ee}{\end{enumerate}}
\newcommand{\beq}{\begin{equation}}
\newcommand{\eeq}{\end{equation}}
\newcommand{\barr}{\begin{array}}
\newcommand{\earr}{\end{array}}
\newcommand{\ii}{\item}
\newcommand{\bi}{\begin{itemize}}
\newcommand{\ei}{\end{itemize}}

\newtheorem{dfn}{Definition}
\newcommand{\bdf}{\begin{dfn}\rm}  
\newcommand{\edf}{\end{dfn}}  
\newtheorem{thm}{Theorem}
\newcommand{\bth}{\begin{thm}\rm}  
\newcommand{\eth}{\end{thm}}  
\newtheorem{exm}{Example}
\newcommand{\bex}{\begin{exm}\rm}  
\newcommand{\eex}{\end{exm}}  
\newcommand{\bpr}{\begin{proof}}  
\newcommand{\epr}{\end{proof}}  
\newtheorem*{hyp}{Hypothesis}
\newcommand{\bhp}{\begin{hyp}\rm}  
\newcommand{\ehp}{\end{hyp}}
\newtheorem*{ppri}{Physical Existence Principle}
\newcommand{\bpri}{\begin{ppri}\rm}
\newcommand{\epri}{\end{ppri}}
\newtheorem*{fpri}{Logical Existence Principle}
\newcommand{\bfpri}{\begin{fpri}\rm}
\newcommand{\efpri}{\end{fpri}}
\newtheorem*{gppri}{General Physical Existence Principle}
\newcommand{\bgpri}{\begin{gppri}\rm}
\newcommand{\egpri}{\end{gppri}}
\newtheorem*{glpri}{General Logical Existence Principle}
\newcommand{\bglpri}{\begin{glpri}\rm}
\newcommand{\eglpri}{\end{glpri}}
\newtheorem*{dpri}{Development Principle}
\newcommand{\bdpri}{\begin{dpri}\rm}
\newcommand{\edpri}{\end{dpri}}

\begin{document} %

\maketitle
\begin{abstract}
A fundamental question is whether Turing machines can model all reasoning processes. 
We introduce an existence principle stating that the perception of the physical existence
of any Turing program can serve as a physical causation for the application
of any Turing-computable function to this Turing program. The existence principle
overcomes the limitation of the outputs of Turing machines to lists, that is, recursively enumerable sets. 
The principle is illustrated by productive partial functions 
for productive sets such as the set of the G\"odel numbers of the Turing-computable total functions. 
The existence principle and productive functions imply the existence of physical systems 
whose reasoning processes cannot be modeled by Turing machines. These systems are called creative.
Creative systems can prove the undecidable formula in G\"odel's theorem in another formal system
which is constructed at a later point in time.
A hypothesis about creative systems, which is based on computer experiments, is introduced. 
\end{abstract}

\section{Introduction}  \label{SIntroduction}

Turing \shortcite[p.\ 21, and 1948, p.\ 17, in the original typescript]{Turing48} discusses the development of intelligence in man and in 
machines:
\bqu
If the untrained infant’s mind is to become an intelligent one,
it must acquire both discipline and initiative. So far we have
been considering only discipline. To convert a brain or machine
into a universal machine is the extremest form of discipline. But
discipline is certainly not enough in itself to produce intelligence.
That which is required in addition we call initiative. ... Our task
is to discover the nature of this residue as it occurs in man, and
to try and copy it in machines.
\equ
Thus, Turing's discipline is the execution of a universal [Turing] machine, that is, the execution of an ordinary computer program. He writes: ``That which is required in addition [to produce intelligence] we call initiative." This means that Turing assumes that intelligence cannot completely be represented by any Turing machine, that is, any computer program, and requires something that is called 
initiative by Turing.

Russell and Norvig \shortcite[p.\ 1020]{Russell...10} refer to an assertion in
 the field of Artificial Intelligence :
\bqu
The proposal for the 1956 summer workshop that defined the field of Artificial Intelligence
(McCarthy {\it et al.}, \citeyear{McCarthy...55}) made the assertion that ``Every aspect of learning or any other feature
of intelligence can be so precisely described that a machine can be made to simulate it.
\equ
Thus, McCarthy {\it et al.} \shortcite{McCarthy...55} assume that ``every aspect of learning or any other feature
of intelligence" can be formalized, that is, can be represented by a Turing machine. In contrast, Turing \shortcite{Turing48} assumes that something %
``is required in addition" to produce intelligence.

The reason why something ``is required in addition" can be illustrated by productive functions
which can be regarded as a formal abstraction of the construction of the undecidable formula in 
G\"odel's \shortcite{Goedel31} incompleteness theorem.

Let $P_1$, $P_2$, $P_3$, ... be a fixed listing of all Turing programs, that is, the set of instructions of all Turing machines (see Rogers \shortcite[p.\ 21]{Rogers87}). The indices 1, 2, 3, ... of
the Turing programs $P_1$, $P_2$, $P_3$, ... 
are called {\em G\"odel numbers}. We write $\varphi_i$ for the partial\footnote{A function is called {\em partial} if it is defined for some  but not necessarily all natural numbers in its input.}  function computed by the
Turing program $P_i$, where $i$ is any G\"odel number.
A Turing-computable partial
function $\psi$ is called {\em productive} for a set $A$ of natural numbers %
if, given any Turing-computable total\footnote{A function is called {\em total} if it it defined for all natural numbers in its input.} 
function $\varphi_i$ 
 whose output
is a subset $S$ of $A$, then
$\psi$ is defined for the input $i$ and the output $\psi(i)$ is contained in $A$, that is,
$\psi(i) \in A$,
 but %
 not %
 in the output
$S$ of $\varphi_i$, that is, 
$\psi(i) \notin S$.\footnote{This definition of productive functions and sets is equivalent to Rogers
\shortcite[pp.\ 84, 90]{Rogers87} because of basic theorems such as Rogers
\shortcite[p.\ 60, Theorem V, and p.\ 61, Corollary V(b)]{Rogers87}.}
Roughly speaking, Turing programs cannot  generate all members
of a productive set $A$
because, given any Turing-computable total function $\varphi_i$ whose output is a subset  $S$ of $A$, then 
$\psi(i) \in A$ and  $\psi(i) \notin S$.
An example of a productive set is the set of the G\"odel numbers of the
Turing-computable total functions whose inputs and output are natural numbers
(see Rogers \shortcite[p.\ 84, {\it Example} 2]{Rogers87}).

A set is called {\em recursively enumerable} if it is empty or the output 
\beq
\{ \varphi_i (1), \varphi_i (2), \varphi_i (3), ... \} \label{varphi123}
\eeq
of a Turing-computable total function $\varphi_i$
(see Rogers \shortcite[p.\ 58]{Rogers87}).
Productive functions can be used to construct a larger recursively enumerable
subset of a productive set $A$ from any given recursively enumerable subset of $A$.
Referring to his definition of a productive partial function for a productive set \cite[p.\ 84]{Rogers87},
Rogers \shortcite[p.\ 90]{Rogers87} writes:
\bqu
It follows from the definition of productiveness that if a set $A$ is productive,
then there is an effective procedure by which, given any recursively enumerable subset of $A$,
we can get a larger recursively enumerable subset of $A$.
\equ
The core of the construction of larger recursively enumerable subsets of a productive set may be summarized in the informal rule
\beq
\mbox{\em if given } P_i\; \;\mbox{\em then apply } \psi\; \mbox{\em to } i, \label{rulegiven}
\eeq
where $P_i$ is a Turing program computing a total function whose output is a subset $S$
of a productive set $A$ and $\psi$ is a productive function for $A$. The application of 
the productive function $\psi$ in the rule (\ref{rulegiven}) to the G\"odel number $i$
of the Turing program $P_i$ yields a natural number $\psi(i) \in A$ that is not contained
in the output $S$ of $P_i$. From the subset $S$ of $A$ and the natural number $\psi(i)$ we
can get a larger subset $S \cup \{\psi(i)\}$ of $A$. 

It is not possible to use the Turing program $P_i$ and the productive function $\psi$ in (\ref{rulegiven})
 to construct a Turing program, say $P_j$, that computes a total function whose output is a 
 subset of the productive set $A$ and contains the rule
(\ref{rulegiven}) such that $P_j$ generates
the output of any given Turing program $P_i$ in (\ref{rulegiven}) 
and the natural numbers $\psi(i)$ although 
$\psi$ is Turing-computable.  A reason is that
the application of the productive function $\psi$ to the G\"odel number $j$
of the Turing program $P_j$ yields a natural number $\psi(j)$ that is not contained
in the output of $P_j$. This means that the application of the productive function $\psi$ 
to $j$ implies that
such a Turing program $P_j$ cannot be given, that is, it cannot exist.
An explanation is that the application of the productive function $\psi$ 
to the G\"odel number $j$ of $P_j$ cannot be achieved in the Turing program $P_j$
itself. Roughly speaking, $P_j$ cannot refer to itself, that is, its own existence.

But a human
can apply the informal rule (\ref{rulegiven}). 
If any Turing program $P_i$ according to (\ref{rulegiven}) is given, %
a human can apply the productive function $\psi$ to $i$ and use the result $\psi(i)$
to produce a set that is larger than the output of $P_i$ although there is no 
Turing program whose output contains all outputs that a human produces by applying 
the informal rule (\ref{rulegiven}).
Thus, productive 
functions,  
which can be regarded as a formal abstraction of the construction of the undecidable formula in 
G\"odel's \shortcite{Goedel31} incompleteness theorem,
provide an explanation for Turing's \shortcite{Turing48}  assumption 
that intelligence cannot completely be represented by any Turing machine, that is, any computer program, and requires something that is called initiative. 
The application of the informal rule (\ref{rulegiven}) appears as
a dynamical process that cannot be formalized in advance because the Turing program $P_i$ in  (\ref{rulegiven}) need not be given at present but can be given in the future by
applying a productive function. This means that the G\"odel numbers of the 
Turing programs $P_i$ in the input of (\ref{rulegiven}) need not be recursively enumerable.

A scientific theory of the informal rule (\ref{rulegiven}), in particular,
a theory of its technical, that is, physical implementation 
requires the solution of three problems. The {\em first} problem
is the problem of {\em existence} because the word ``{\it given}" in (\ref{rulegiven}) must
refer to any given, that is, existing Turing program $P_i$.
This problem is fundamental because 
the set of the possible G\"odel numbers of the  
Turing programs $P_i$ in the input of (\ref{rulegiven}) is not recursively enumerable.
The {\em second} problem is how a {\em reference} to a ``{\it given}", that is, existing Turing program $P_i$ in (\ref{rulegiven}) can be established. 
The {\em third} problem is the problem of the physical {\em causation}
for the application of the productive function $\psi$ in (\ref{rulegiven}), that is,
the push of the button to apply $\psi$ to $i$.

The three problems of a scientific theory of the informal rule (\ref{rulegiven}),
that is, the problems of the existence of a Turing program $P_i$, the reference
to $P_i$, and the causation to apply $\psi$ to $i$ are solved by an {\em existence
principle} stating that the perception of the physical existence of a Turing program
can serve as a physical causation for the application of any Turing-computable function, for example,
a productive function, to this information.

Section \ref{SExistence}
introduces the existence principle.
Section \ref{SReasoning} outlines implications of the existence principle
for reasoning processes.
Section \ref{SDiscussion} discusses our results, in particular, the Church-Turing thesis.

\section{Existence}\label{SExistence}
The following physical existence principle solves the three problems of a scientific theory
of the informal rule (\ref{rulegiven}) in Section \ref{SIntroduction}, that is,
the problems of the existence of a Turing program $P_i$, the reference
to $P_i$, and the causation to apply a productive function $\psi$ to the G\"odel number $i$
of $P_i$.
\bpri
Let $\psi$ be a productive partial function for a productive set $A$.
The perception of the physical existence of any Turing program $P_i$ 
that computes a total function whose output is a subset of the productive set $A$
can serve as a physical causation for the application of %
the productive function $\psi$ 
to the G\"odel number $i$ of $P_i$.
\epri
The principle solves the problem of the existence of the Turing program $P_i$
in the informal rule (\ref{rulegiven}) because it refers to the physical
existence of $P_i$. 
Therefore, the set of
the G\"odel numbers of the  
Turing programs $P_i$ in the input of (\ref{rulegiven})
need not be recursively enumerable.
Thus, the informal rule (\ref{rulegiven}) can process a set of
G\"odel numbers that is not recursively enumerable because the set of the G\"odel numbers that
satisfy the conditions in (\ref{rulegiven}) is productive.
The physical existence principle also solves the problem
of the reference to $P_i$ because the perception
of the physical existence of %
the Turing program $P_i$ in (\ref{rulegiven})
establishes a reference to $P_i$. %
For example, this perception can be achieved by light, that is,
electromagnetic waves. A human can use his eyes to perceive
a physical representation of $P_i$. %
A technical input device is a camera that is connected to a computer.
In general, this perception can be achieved by any physical means, for example,
the physical means for the perception of a Turing program $P_i$ 
that is represented in the brain of a human.
Finally, the physical existence principle also solves the problem
of causation because the perception of the physical representation
of $P_i$ %
can cause a human to start the execution of a Turing program,
that is, a computer program. 
The perception of the physical representation
of $P_i$ %
by means of a connected camera can cause a computer to start a program. 

\bdf \label{creativesystem}
A physical system that contains an implementation of the physical existence principle is called {\em creative system}.\footnote{Ammon \shortcite[Section 3.1]{Ammon87} defines a creative system
by requiring that it can determine outputs of functions that cannot be computed by Turing
programs, that is, for a given Turing program a creative system can determine 
such outputs that are not contained in the outputs of the Turing program.
Roughly speaking, a creative system cannot be modeled by a Turing program because
it can use the program as a basis for its further development.}
\edf
By means of the physical existence principle the informal rule (\ref{rulegiven}) 
in Section \ref{SIntroduction} can be
transformed into the following precise physical rule:
\br \label{rphysical}
  Let $\psi$ be a productive partial function for a productive set $A$.
  Then 
  \beq
\mbox{\em if } \mbox{\bf\em exists } P_i\; \;\mbox{\em then apply } \psi\; \mbox{\em to } i, \label{ruleexists}
\eeq
where $P_i$ is a Turing program computing a total function whose output is a subset $S$
of the productive set $A$
and
{\bf\em exists} means that $P_i$ exists physically,
describes an implementation of the physical existence principle.
\er
Thus, the word {\bf\em exists} in (\ref{ruleexists}) refers to a physical process,
that is, that the perception of the physical existence of $P_i$
serves as a physical causation for the application of the productive function $\psi$
to the G\"odel number $i$ of $P_i$ according to the physical existence principle.
 
If a creative system according to Definition \ref{creativesystem}
contains an implementation of the physical rule (\ref{ruleexists}),
it applies the physical rule (\ref{ruleexists}), that is,
 it applies  the productive function $\psi$
to the G\"odel number $i$ of any {\em existing} Turing program $P_i$
computing a total function whose output is a subset $S$
of the productive set  $A$ according to (\ref{ruleexists}).
This means that the G\"odel numbers $i$ in the input
of the physical rule (\ref{ruleexists}) need not be recursively enumerable
because $A$ is a productive set
such that for  
any Turing program $P_i$
computing a total function whose output is a subset $S$
of $A$
there {\em exists} 
a Turing program
computing a total function whose output is a subset
of $A$ that is larger that $S$.

If we abstract the logical aspects from the physical existence principle
we get the following principle:
\bfpri
Let $\psi$ be a productive partial function for a productive set $A$.
A creative system can apply  the productive function  $\psi$ %
to the G\"odel number $i$ of any existing Turing program $P_i$
that computes a total function whose output is a subset of the productive set $A$.
\efpri

By means of the logical existence principle we can prove theorems about
creative systems.

Roughly speaking, the following theorem states that 
the outputs of creative systems cannot be generated by Turing programs.
\begin{thm}\label{tphysical}
Let $\psi$ be a productive partial function for a productive set $A$,
let $C$ be a creative system that applies the physical rule 
(\ref{ruleexists}), and let $C$ produce any sequence
\beq
j_1, j_2, j_3, ... \label{tphysicalseq}
\eeq
of G\"odel numbers that are contained in the productive set $A$.
Then,
there exists no Turing program whose output is the sequence (\ref{tphysicalseq}).
\end{thm}
\bpr
Let 
$P_i$ be an existing Turing program
that produces the 
sequence (\ref{tphysicalseq})
of G\"odel numbers.
Because the creative system
 $C$ applies the physical rule (\ref{ruleexists}),
 $C$ produces an output $\psi(i)$ by applying the productive function $\psi$ to the G\"odel number $i$ of $P_i$.
 Because the G\"odel numbers $j_1$, $j_2$, $j_3$, ... in (\ref{tphysicalseq}) are contained in 
 the productive set $A$ and $\psi$ is a productive function for $A$,
 $\psi(i)$ is not contained in the output of $P_i$, that is, $\psi(i)$ is different from all
 G\"odel numbers $j_1$, $j_2$, $j_3$, ... in (\ref{tphysicalseq}).
 Therefore, the Turing program $P_i$ does not produce
 the sequence (\ref{tphysicalseq}) which is produced by the creative system $C$.
 Thus, our original assumption that the Turing program $P_i$ produces the 
sequence (\ref{tphysicalseq}) yields a contradiction. 
Hence, there exists no Turing program whose output is the sequence (\ref{tphysicalseq}).
\epr
Theorem \ref{tphysical} implies the existence of creative physical systems
that cannot be modeled by any Turing program, that is, Turing machine,
because creative systems can perceive the physical existence of the Turing machine
and thus use this machine as a basis to 
produce an output that is not contained in the output of this machine.

\section{Reasoning}\label{SReasoning}

The Turing program $P_i$ in input of the physical rule (\ref{ruleexists}) 
in Section \ref{SExistence} can be interpreted
as a proposed description of the sequence (\ref{tphysicalseq}) in Section \ref{SExistence}
which is produced by
the creative system $C$ in Theorem \ref{tphysical}. 
$C$ applies the productive function $\psi$
in (\ref{ruleexists}) to the G\"odel number  $i$ of $P_i$, that is, to the proposed description
of the sequence (\ref{tphysicalseq}).
This produces an output $\psi(i)$ which is not
contained in the output of $P_i$ according to the proof of Theorem \ref{tphysical}.
The output of $P_i$ is a subset of the productive set $A$ in Theorem \ref{tphysical}.
The output $\psi(i)$ of $C$ and the output 
\beq
\{\varphi_i(1), \varphi_i(2), \varphi_i(3), ... \} \label{outputpi}
\eeq
of $P_i$, where $\varphi_i$ is the function computed by $P_i$,
form a larger 
recursively enumerable subset 
\beq
\{\psi(i), \varphi_i(1), \varphi_i(2), \varphi_i(3), ... \} \label{outputpj}
\eeq
of $A$ which is the output of another Turing
program, say $P_j$.
Thus, the creative system $C$ uses a Turing program $P_i$ in its input, which
can be regarded as a proposed description of its output (\ref{tphysicalseq}), to produce an output 
$\psi(i)$ that can be used to construct a Turing program $P_j$ whose output 
(\ref{outputpj}) is
larger than the output (\ref{outputpi}) of $P_i$.

The construction of a Turing program $P_j$ whose output (\ref{outputpj}) is
larger than the output (\ref{outputpi}) of any $P_i$ that 
exists physically and computes a total function whose output is subset of a productive set cannot be formalized because there is no general formal procedure for the application of
the productive function $\psi$ to
the G\"odel number $i$ within $P_i$.
But the physical rule (\ref{ruleexists}), which is an implementation of the 
existence principle, applies $\psi$ to
the G\"odel number $i$ of any $P_i$ that satisfies the conditions given above such that
the output $\psi(i)$ can be used to construct a Turing program $P_j$ whose
output (\ref{outputpj}) is larger than the output (\ref{outputpi}) of $P_i$.

The G\"odel number $i$ of the Turing program $P_i$ in 
the physical rule (\ref{ruleexists}), which is an implementation of the 
existence principle, refers to $P_i$ {\em as a whole}. This holistic aspect 
of the physical rule (\ref{ruleexists}) explains why 
there is no general formal procedure for the application of
the productive function $\psi$ to
the G\"odel number $i$ within $P_i$.
Productive functions such as the function $\psi$ 
in the physical rule (\ref{ruleexists})
can be regarded as a formal abstraction of the construction of the undecidable formula in 
G\"odel's \shortcite{Goedel31} incompleteness theorem.
Kleene \shortcite[p.\ 426]{Kleene52} writes:
 ``we can recognize that  [the undecidable formula] ${\rm A}_p(\boldsymbol{p})$ 
 [in the number-theoretic formal system]
is true by taking into view the structure of that system as a whole".
Thus, the recognition of the truth of the  undecidable formula ${\rm A}_p(\boldsymbol{p})$ 
in
G\"odel's theorem, which cannot be proved in the number-theoretic formal system according 
to G\"odel's theorem,  requires a reference to the (incomplete) formal ``system
as a whole"
which cannot be achieved within the formal system itself
because the formal system  cannot take ``into view the structure of that system as a 
whole".
In particular, the Turing program $P_i$ 
in the physical rule (\ref{ruleexists})
is a formal system which cannot take ``into view the structure of that system as a whole"
because the formal system, that is, the Turing program $P_i$
 cannot %
 use its own G\"odel number $i$ to produce the
  result $\psi(i)$ of applying the productive partial function $\psi$
for the productive set $A$ to $i$,
that is, the application of $\psi$ to the G\"odel number $i$ of $P_i$
cannot be achieved within $P_i$.
But we can recognize the truth that $\psi(i)$ is a member of the productive set $A$.
An explanation for our capability to recognize that
$\psi(i)$ is a member of $A$ and the impossibility 
of the Turing program $P_i$ to apply $\psi$ to the G\"odel number $i$ of $P_i$
within $P_i$ can be found in restrictions for formal systems.
Kleene \shortcite[p.\ 64]{Kleene52} 
writes: 
\bqu
Metamathematics must study the
formal system as a system of symbols, etc.\ which are considered wholly objectively. This means 
simply that those
symbols, etc.\ are themselves the ultimate objects, and are not being used to refer to something other than
themselves. The metamathematician looks at them, not through and beyond them; thus they are objects without interpretation or meaning.
\equ
Thus, a reference of a formal system to itself as a whole
cannot be achieved in the formal system,
for example, the use of the G\"odel number $i$
of the Turing program $P_i$  in 
 (\ref{ruleexists}) within $P_i$ itself.
The physical existence principle describes a general reference   
to a formal system as a whole because the perception of 
the physical existence of the Turing program $P_i$, that is, the formal system, establishes
a general reference to this formal system as a whole.
For example, in the physical rule  (\ref{ruleexists}) the G\"odel number $i$
is such a general reference to $P_i$ as a whole.

Let $\psi$ be a productive partial function for a productive set $A$,
let $P_x$ be a physically existing Turing program that computes a total function whose output is a subset of $A$, and
let $C$ be a creative system that applies the physical rule 
(\ref{ruleexists}) in Section \ref{SExistence}. 
We assume that the G\"odel number $x$ of $P_x$ is not known.
The creative system $C$ can find the G\"odel number $x$
of $P_x$  by generating $1$, $2$, $3$, ..., $i$, ... and comparing 
$P_1$, $P_2$, $P_3$, ..., $P_i$, ... with $P_x$. If $P_i$ is identical with $P_x$,
then the creative system $C$ applies  physical rule 
(\ref{ruleexists}), that is, it produces an output $\psi(i)$
by applying $\psi$ to the  G\"odel number $i$ of $P_i$
which is identical with the Turing program $P_x$ that exists physically.
If the Turing program  $P_x$ computes a total function whose output is a 
subset of $A$, then  $\psi(i)$ is not in the output of $P_x$, that is, $P_i$,
according to Theorem \ref{tphysical} in Section \ref{SExistence}.
Thus, creative systems can take
 ``into view the structure of [the Turing program $P_x$, that is, $P_i$] as a whole"
 when they determine the G\"odel number $i$ of a physically existing Turing program $P_x$
 whose G\"odel number $x$ is not known.
 This implies that creative systems can produce the output $\psi(i)$ by applying $\psi$ to the G\"odel number $i$
 of any physically existing Turing program $P_i$
  computing a total function whose output is a subset of $A$ although 
  there is no general formal procedure to achieve this within $P_i$.

If we assume that the child's mind can be represented by a formal system, that is, a Turing program,
at every point in time
the existence principle implies that the core of Turing's \shortcite{Turing48} initiative in Section \ref{SIntroduction}
is an informal physical reference of the child's mind to itself as a whole at any point
in time.
This reference is established by the perception of the physical
structures in the child's mind which form the basis for their further development, that is,
there is no formal description of the child's development that can be given in advance
but the formal description at any point in time is the basis and method of its
further development.
Thus, the existence principle implies that the core of the development of the child's mind,
that is, the core of intelligence, is an informal physical reference and application of
the child's mind to itself as a whole at any point in time.

\section{Discussion}\label{SDiscussion}

Referring to 
his ``Theorem 2.4, with its corollaries"
Davis \shortcite[pp.\ 121-122]{Davis58} writes:
\bqu
... these results 
really constitute an abstract form of G\"odel's famous incompleteness theorem ... they imply that 
{\em an adequate development of the theory of natural numbers, within a logic L, to the point 
where membership in some given set $Q$ of integers can be adequately dealt with
within the logic ... is possible only if $Q$ happens to be recursively enumerable.}
Hence, non-recursively enumerable sets can, at best, be dealt with in an incomplete manner.
\equ
This implies that the sequence (\ref{tphysicalseq}) in Theorem \ref{tphysical}
in Section \ref{SExistence},
which is produced by a creative system $C$ by means of the physical rule (\ref{ruleexists}),
cannot be dealt with within a logic L because 
the sequence (\ref{tphysicalseq}) is not recursively enumerable
according to Theorem \ref{tphysical},
that is, there is no Turing program, say $P_i$,
whose output is (\ref{tphysicalseq}).
The output of $P_i$ corresponds to the 
``given set $Q$ of integers" in the above quotation from
Davis \shortcite[pp.\ 121-122]{Davis58}.
In the proof of Theorem \ref{tphysical} 
the creative system $C$ applies the 
productive function $\psi$ to the
G\"odel number $i$ of $P_i$ which produces an output $\psi(i)$
that is not contained in the output of $P_i$ but
in the productive set $A$.
Roughly speaking, creative systems can
overcome the limits of any logic that deals with subsets of productive sets because they
can apply productive functions, which can be computed by Turing machines,
to the recursively enumerable sets to which the logic is restricted.

Turing's thesis states that every function that would be naturally regarded as computable
is computable under his definition, that is, by one of his machines \cite[pp.\ 376-381]{Kleene52}.	
Turing  \shortcite[p.\ 231]{Turing36} restricts his machines to a finite number of {\em m}-configurations (machine configurations)
which are called ``states of mind" 
in %
his thesis.
Turing \shortcite[pp.\ 249-250]{Turing36} 
supposes that the ``number of states of mind" is finite
because some of them ``will be confused" if ``we admitted an infinity of states of mind"
(see Kleene \shortcite[pp.\ 376-377]{Kleene52}). 
G\"odel \shortcite[p.\ 306]{Goedel72a}
regards the restriction to a finite number of states as a ``philosophical error
in Turing's work"
and points out
that ``mental procedures" may ``go beyond mechanical procedures".
G\"odel \shortcite[p.\ 306]{Goedel72a} writes:
\bqu
What Turing disregards completely is the fact that
{\em mind, in its use, is not static, but constantly developing}, ... 
There may exist systematic methods of actualizing this development, which could form part of the procedure.
Therefore, 
Turing's number of {\em distinguishable states of mind} %
may {\em converge toward infinity}
 in the course of the application of the procedure.
\equ
The Turing program $P_i$ in Section \ref{SExistence}
produces an output (\ref{outputpi}).
The Turing program $P_j$ in Section \ref{SExistence}
produces a larger output (\ref{outputpj}).
The program $P_j$ can be constructed from the Turing program $P_i$
and the productive function $\psi$ by means of the 
 physical rule  (\ref{ruleexists}) in Section \ref{SExistence}.
 The 
 physical rule  (\ref{ruleexists}), which is an implementation of the physical existence
 principle, can be regarded
 as a ``systematic method" that cannot be modeled by any Turing machine 
 according to Theorem \ref{tphysical}.
 In view of the structure of its output (\ref{outputpj}) the program $P_j$ can 
 be constructed from the Turing program $P_i$
and the productive function $\psi$, which is computed by a Turing program, say $P_k$, in a straightforward
manner such that the number of states in $P_j$ is greater than the sum 
 of the number of states in $P_i$ and $P_k$.
 This suggests that the repeated application of the 
 physical rule  (\ref{ruleexists}) can produce Turing programs
 whose number of states grows in the  course of time.
 Thus, Turing's assumption that the number of states is finite or infinite
 is misleading because the number of states may grow in the course of time.
 Therefore, a concept of time that cannot be formalized is necessary to understand
 the development of ``mind".

Ammon \shortcite{Ammon93} describes an automatic proof of G\"odel's incompleteness
theorem by a system that is composed of heuristics.
Because these heuristics are rather elementary
they might be constructed automatically.
Such computer experiments (see Ammon \shortcite{Ammon88}, \shortcite{Ammon92a}, and
\shortcite{Ammon92b}) and our theoretical results
suggest the following principle:
\bdpri
A creative system is a self-developing process which starts from any universal
programming language and any input. This process cannot be reduced
to a Turing machine but to the language and the input from which it starts. 
\edpri

Thus, the structures in a creative system that exist presently
form the basis, that is, the input and the method for its further development.
This cannot be achieved by Turing programs.
For example, the application of the productive function $\psi$
to the G\"odel number $i$ of a Turing program $P_i$
in the  physical rule  (\ref{ruleexists})
in Section \ref{SExistence}
cannot be achieved within $P_i$.
The development principle implies that
a creative system can be represented as a finite sequence of 0s and 1s or
a binary number at any point in time but it can construct and change 
any of its structures in the course of time.\footnote{Referring to 
computer experiments Ammon \shortcite[Section 5.3]{Ammon87} writes:
``The experiments show that the knowledge required for mathematical research 
is surprisingly simple and domain specific, but subject to rapid modifications
and extensions."}
This has implications for the technical development of
a creative system
because it should start with the manual development of a rather simple
domain-specific language for a user interface which can be used
for its further development.

The proof of Theorem \ref{tphysical} in Section \ref{SExistence}
derives a contradiction from the assumption 
that an {\em existing} Turing program $P_i$ computes the output (\ref{tphysicalseq}) of 
a creative system $C$. The proof uses the G\"odel number $i$ of $P_i$ which
refers to $P_i$ as a {\em whole}.
The %
reference of %
$i$ to $P_i$ %
is contained in the physical rule (\ref{rphysical}) in the proof
of Theorem \ref{tphysical}. 
This physical
reference of the G\"odel number $i$ to $P_i$ cannot be modeled by any
Turing machine $P_i$ according to the proof %
because the
application of the productive function $\psi$ to the G\"odel
number $i$ cannot be achieved within $P_i$, that is, there is no {\em general\/}
Turing program modeling the application of $\psi$ to the G\"odel
number $i$.

Although the {\em physical\/} reference of the G\"odel number $i$ to an {\em existing} $P_i$
in the proof
of Theorem \ref{tphysical}
is not Turing-computable, the construction of $\psi(i)$, which is not in the output of
$P_i$, can be formally represented in another more powerful Turing program
that depends on the G\"odel number $i$ of $P_i$.
For example, such a more powerful Turing program
is the Turing program %
that produces the extended sequence
(\ref{outputpj}) in Section \ref{SReasoning}.
Even by means of the physical rule (\ref{rphysical}), that 
is, the physical existence principle, we cannot construct a general Turing program
because we could apply a productive function $\psi$ to such a Turing program
according to  Theorem \ref{tphysical}, that is, there is {\em no such  general} Turing program.

The application of the physical existence principle, for example the physical rule (\ref{rphysical}),
requires resources, in particular {\em time}. 
The Turing program $P_i$ in the physical rule (\ref{rphysical}) must exist physically before
it can be perceived according to the physical existence principle. 
The output $\psi(i)$ of the application of $\psi$ to $i$ and the extended sequence
(\ref{outputpj}) can exist only at a later point in time.
According to  Theorem \ref{tphysical} the output $\psi(i)$ and the extended sequence
(\ref{outputpj}) cannot be described in advance by a general formal system, that is, a Turing program.
By means of the physical existence principle, for example the physical rule (\ref{rphysical}),
more and more powerful Turing programs can be constructed in the course of time.
As soon as the
output $\psi(i)$ or the extended sequence
(\ref{outputpj})
 exist physically and have been perceived according
to the physical existence principle
their construction can be described formally. Therefore, an informal concept of time is 
a prerequisite for an understanding of the informal physical evolution
of formal systems such as Turing programs.

The physical existence principle and the logical existence principle in Section \ref{SExistence}
are restricted
to any productive partial function $\psi$ for a productive set $A$
and any Turing program $P_i$ 
that computes a total function whose output is a subset of $A$.
We can generalize the physical  existence principle and the logical existence principle 
by allowing any Turing-computable partial function $\psi$
and any Turing program $P_i$ that satisfies any %
properties, that is,
properties that cannot be formalized in any single formal system.
\bgpri \label{gpri}
Let $\psi$ be a Turing-computable partial function.
The perception of the physical existence of any Turing program $P_i$ 
that satisfies any %
properties
can serve as a physical causation for the application of %
the function $\psi$ 
to the G\"odel number $i$ of $P_i$.
\egpri
\bglpri \label{glpri}
Let $\psi$ be a Turing-computable partial function.
A creative system can apply  the function  $\psi$ %
to the G\"odel number $i$ of any existing Turing program $P_i$
that satisfies any %
properties.
\eglpri
By means of this generalized existence principle a creative system 
``can recognize that [the undecidable formula] ${\rm A}_p(\boldsymbol{p})$ 
[in G\"odel's incompleteness theorem]
is true
by taking into view the structure of that [number-theoretic formal] system as a whole" (see 
 Kleene, \shortcite[p.\ 426]{Kleene52}),
that is, it can overcome the incompleteness of the formal system.
Such a formal system can be represented as a Turing program $P_i$
satisfying some simple properties, in particular, the property that the formal system is consistent.
This means that a Turing-computable partial function $\psi$ produces Kleene's 
 \shortcite[p.\ 426]{Kleene52} representation of the 
undecidable formula 
 ${\rm A}_p(\boldsymbol{p})$ 
 in G\"odel's theorem from the G\"odel number $i$ of the 
 Turing program $P_i$. %
 A formal proof of G\"odel's  
 undecidable formula 
 ${\rm A}_p(\boldsymbol{p})$ 
 can be achieved in {\em another} formal system, say $S_i$,
that depends on the G\"odel number $i$ of $P_i$
representing the (incomplete)
 number-theoretic formal system. %
 This implies that a creative system can {\em prove} the truth of 
G\"odel's  
 undecidable formula 
 ${\rm A}_p(\boldsymbol{p})$ in the formal system $S_i$ because it can refer to the 
 (incomplete) number-theoretic formal system as a whole.
 Such a proof depends on the prerequisite that
 this number-theoretic formal system is consistent (see Kleene
 \shortcite[p.\ 426]{Kleene52}).
Because the application of the physical existence principle
requires resources, in particular {\em time}, 
the Turing program $P_i$, which represents the number-theoretic formal system, must exist physically before
it can be perceived according to the physical existence principle. 
The output $\psi(i)$ of the application of $\psi$ to the G\"odel number $i$ and 
the proof of $\psi(i)$, that is, the undecidable formula
${\rm A}_p(\boldsymbol{p})$ in the formal system $S_i$ 
 can exist only at a later point in time.
 According to G\"odel's theorem the output $\psi(i)$, 
 that is, the undecidable formula
${\rm A}_p(\boldsymbol{p})$ cannot be proved in 
the number-theoretic formal system, which is represented by $P_i$, but
it can be proved by a creative system at a later point in time in another formal system $S_i$ 
which can be constructed by 
means of the existence principle. This applies to any existing 
number-theoretic formal system.
By a repeated application of the physical existence principle a creative system can construct
more and more powerful formal systems in the course of time
which cannot be described by any (existing) formal system in advance
although any individual construction can be proved in a formal system $S_i$
with the benefit of hindsight.
This confirms that an informal concept of time is 
a prerequisite for an understanding of the informal physical evolution
of formal systems which can be represented by Turing programs.
We discuss the application of the general existence principle
to G\"odel's incompleteness theorem in detail in the following paragraphs. 

Davis \shortcite[p.\ 611]{Davis93} argues that G\"odel's incompleteness theorem
can be proved in a formal system:
\bqu
It [G\"odel's theorem] is, however, a quite ordinary sentence of elementary number theory and can be proved with no
difficulty whatever in any formal system adequate for elementary number theory, such as
for example Peano arithmetic.
Note that this powerful form of G\"odel's theorem applies
uniformly to any formalism whatever.
\equ
A formal system, say $S_1$, %
``adequate for elementary number theory, such as
for example Peano arithmetic", in which G\"odel's theorem is proved, is incomplete as well. 
Thus, a proof of G\"odel's theorem for
$S_1$ %
requires another formal system, say $S_2$, which refers to $S_1$. %
A proof of G\"odel's theorem for $S_2$ requires another formal system,
say $S_3$, which refers to $S_2$, and so on. Thus, there is no proof in any formal
system showing that G\"odel's theorem applies ``to any formalism whatever".
An exception is any formal system itself such as 
$S_1$, %
$S_2$, $S_3$, and so on 
in which G\"odel's theorem is proved.

The first part of G\"odel's (first incompleteness) theorem states that the undecidable 
formula ${\rm A}_p(\boldsymbol{p})$ is unprovable in the formal system
if the system is consistent  (see Kleene \shortcite[p.\ 207, Theorem 28]{Kleene52}).
The following definition  
prepares a theorem stating
that a formalization of
the proposition that \mbox{\App} is true can be proved in another formal system
that is called {\em observing system}. Roughly speaking, an {\em observing system}
formalizes the view of a mathematician who proves the first part of G\"odel's theorem.
 
\bdf \label{observing}
Let $S$ be a formal system. An {\em observing system} of 
the {\em observed system} $S$ is a formal system $\overline{\mbox{$S$}}$
that is a copy of $S$ containing a formal symbol $\boldsymbol{S}$ for $S$.
If $F$ is a formal expression, that is, 
a finite sequence of formal symbols, in $S$,
we write $\overline{\mbox{$F$}}$ for the formal expression that represents $F$ in $\overline{\mbox{$S$}}$, that is, it contains the formal symbol  $\boldsymbol{S}$  %
for $S$. %
 $\overline{\mbox{$F$}}$ is called the {\em observing formula} of the 
 {\em observed formula} $F$ in $S$.
\edf
 
\bex \label{Apstrokepstroke}
The construction of undecidable formula ${\rm A}_p(\boldsymbol{p})$ in G\"odel's theorem
is based on a predicate $A(a,b)$ which is defined by  (see Kleene \shortcite[p.~206, Lemma 21]{Kleene52}): 

$A(a,b)$: $a$ is the G\"odel number of a formula ${\rm A}_a(\textsl{a})$ with a free variable 
$\textsl{a}$ and $b$ is the G\"odel number of a proof of the formula ${\rm A}_a(\boldsymbol{a})$, where
$\boldsymbol{a}$ is the formal expression representing the number $a$.\footnote{We write
${\rm A}_n$ for the formula whose G\"odel number is $n$.
For ${\rm A}_n$ we may write
${\rm A}_n(\textsl{a})$ showing the free variable $\textsl{a}$ 
for use with substitution (see Kleene \shortcite[p.~206, Lemma 21]{Kleene52}).}

The definition of $A(a,b)$ implicitly refers to formulas in a formal system, say, $S$.
We extend the definition of $A(a,b)$ by including an explicit reference to $S$.
This yields:

$A'(a,b)$: $a$ is the G\"odel number of a formula ${\rm A}_a(\textsl{a})$ in $S$ with a free variable $\textsl{a}$
and $b$ is the G\"odel number of a proof of the formula ${\rm A}_a(\boldsymbol{a})$
in $S$, where
$\boldsymbol{a}$ is the formal expression representing the number $a$.

The predicate $A(a,b)$ can be expressed by a formula ${\rm A}(\textsl{a},\textsl{b})$ in $S$ (see Kleene \shortcite[p.~207]{Kleene52}).
Analogously, the extended predicate $A'(a,b)$ can be expressed by
a formula  ${\rm A'}(\textsl{a},\textsl{b})$ in the formal system $\overline{\mbox{$S$}}$
that is an observing system of $S$ and
contains a formal symbol  $\boldsymbol{S}$
for $S$ (see Definition \ref{observing}).
Thus, the formula  ${\rm A'}(\textsl{a},\textsl{b})$ is the observing formula  $\overline{\mbox{${\rm A}(\textsl{a},\textsl{b})$}}$
of the formula  ${\rm A}(\textsl{a},\textsl{b})$ in $S$.
Analogously to the construction of the undecidable formula \mbox{\App},
that is, 
 \beq
\forall {\rm \textsl{b}} \, \neg {\rm A}(\boldsymbol{p}, \textsl{b}), \label{Appforall}
\eeq
from the formula %
${\rm A}(\textsl{a},\textsl{b})$ in $S$
(see Kleene
\shortcite[p.~207]{Kleene52}),
 we can construct an observing formula 
$\overline{\mbox{\App}}$ from the formula %
${\rm A'}(\textsl{a},\textsl{b})$ in 
$\overline{\mbox{$S$}}$ as follows: 
Let $p'$ be the G\"odel number of the formula 
$\forall {\rm \textsl{b}} \, \neg {\rm A'}(\textsl{a}, \textsl{b})$,
that is, ${\rm A}_{p'}(\textsl{a})$,
which contains the free variable \textsl{a} and no other free variable.
The substitution of the formal representation
$\boldsymbol{p'}$ of the G\"odel number $p'$ for the variable \textsl{a}
in ${\rm A}_{p'}(\textsl{a})$, which uses Cantor's diagonal method, yields 
\Apstrokepstroke,
that is, 
\beq
\forall {\rm \textsl{b}} \, \neg {\rm A'}(\boldsymbol{p'}, \textsl{b}), 
\eeq
which is the observing formula $\overline{\mbox{\App}}$ 
of G\"odel's undecidable formula 
 \mbox{\App} 
 (see Kleene \shortcite[p.~207]{Kleene52}).

The following theorem states
that a formalization of
the proposition that the undecidable formula \mbox{\App} 
in an observed formal system is true can be proved in the observing formal system.
\eex 
 
\begin{thm} \label{tprovable}
Let \mbox{\App} be the undecidable formula in a formal system $S$.
The observing formula  $\overline{\mbox{\App}}$ of 
\mbox{\App} is provable 
in the observing system $\overline{\mbox{$S$}}$ of $S$ if $S$ is consistent.
\end{thm}
\bpr
The first half of G\"odel's theorem states that the 
formula \mbox{\App} is unprovable in the formal system $S$ 
if $S$ is consistent  (see Kleene \shortcite[p.~207, Theorem 28]{Kleene52}).
Because the proposition that \App\ is unprovable in $S$ 
is expressed, via the G\"odel numbering, by the formula \App,
\App\ is a formalization of 
the proposition 
that
 \App\ is unprovable in $S$.\footnote{Kleene \shortcite[p. 207]{Kleene52} writes: 
``.. we can interpret the formula  ${\rm A}_p(\boldsymbol{p})$ from our perspective of G\"odel numbering as expressing the proposition that ${\rm A}_p(\boldsymbol{p})$  is unprovable, i.e., 
it is a formula A which asserts its own unprovability.}
 The proposition that $S$ is consistent can be formalized as well
 (see Kleene \shortcite[p.\ 210]{Kleene52}).
 Let {\em Consys} be a formalization of the proposition that $S$ is consistent.
 The proof of the first half of G\"odel's theorem that the 
formula \App\ is unprovable in $S$
if $S$ is consistent can also be formalized in $S$ 
(see Kleene \shortcite[pp.\ 210-211]{Kleene52}).
Thus, we have
\beq
 \vdash_{S} \mbox{\em Consys} \implies \mbox{\App}, \label{pGtS0}
\eeq 
that is, there is a proof in $S$ that the consistency of $S$ implies the formula 
${\rm A}_p(\boldsymbol{p})$ in $S$.\footnote{This proof is a part
of the proof of G\"odel's second incompleteness theorem
which states that 
the consistency of $S$ cannot be proved in $S$ (see Kleene \shortcite[pp. 210-211]{Kleene52}).}
The formula 
\beq
\mbox{\em Consys} \implies \mbox{\App} \label{pGtS0i}
\eeq
in (\ref{pGtS0}) corresponds to
the observing formula
\beq
\overline{\mbox{\em Consys}\vphantom{)}} \;\; %
\mbox{$\Longrightarrow$} \;\; \overline{\mbox{\App}}, \label{pGtS1i}
\eeq
in $\overline{\mbox{$S$}}$
and the proof of (\ref{pGtS0i}) in $S$ %
corresponds to the observing proof
of (\ref{pGtS1i})
in $\overline{\mbox{$S$}}$.
Thus, 
\beq
\vdash_{\overline{S}} \overline{\mbox{\em Consys}\vphantom{)}} \;\; %
\mbox{$\Longrightarrow$} \;\; \overline{\mbox{\App}}, \label{pGtS1}
\eeq
that is, there is a proof in $\overline{\mbox{$S$}}$ that the consistency of $S$ implies the observing formula 
$\overline{\mbox{\App}}$ of \mbox{\App}.
Therefore, (\ref{pGtS1}) corresponds to the theorem.
Thus, the proof is complete.
\epr

The observing formula 
$\overline{\mbox{\App}}$ in $\overline{\mbox{$S$}}$
corresponds to the formula \Apstrokepstroke\
in Example \ref{Apstrokepstroke}.
Because $\overline{\mbox{$S$}}$ is a copy of $S$
containing a formal symbol for $S$,
the observed theorems of all theorems in $S$ can
be proved in $\overline{\mbox{$S$}}$.
Because  the 
observing formula 
$\overline{\mbox{\App}}$
of the undecidable formula \App, which is not provable
in $S$ according to G\"odel's theorem,
is provable in $\overline{\mbox{$S$}}$,
the representation of the 
formal expressions in $\overline{\mbox{$S$}}$
by G\"odel numbers in the formal system $S$
yields proofs of more theorems.
Roughly speaking, the observing formal system
 $\overline{\mbox{$S$}}$ is more powerful than 
 the observed formal system $S$.

The observing formula 
$\overline{\mbox{\App}}$, which
contains a formal symbol $\boldsymbol{S}$ for $S$,
 can be interpreted as a formalization of
the proposition that \mbox{\App} is true from 
a mathematician's point of view
who refers to $S$, 
that is, the observing system $\overline{\mbox{$S$}}$
is a formalization of 
a mathematician's point of view who
 proves G\"odel's theorem.
 The formal system $\overline{\mbox{$S$}}$ depends on $S$, that is, it can be constructed from $S$
 because it is a copy of $S$ containing a formal 
 symbol $\boldsymbol{S}$ for $S$ (see Definition \ref{observing}).

Referring to his undecidable formula ${\rm A}_p(\boldsymbol{p})$ in G\"odel's 
incompleteness theorem 
Kleene \shortcite[p.\ 426]{Kleene52} writes:
\bqu
... if we suppose the number-theoretic formal system
to be consistent, we can recognize that ${\rm A}_p(\boldsymbol{p})$ 
is true by taking into view the structure of that system as a whole,
though
we cannot recognize the truth of  ${\rm A}_p(\boldsymbol{p})$ 
by use only of the principles of inference formalized in that system,
i.e. not $ \vdash {\rm A}_p(\boldsymbol{p})$.\footnote{The expression
``not $ \vdash  {\rm A}_p(\boldsymbol{p})$" %
in  Kleene \shortcite{Kleene52} means
that the undecidable formula ${\rm A}_p(\boldsymbol{p})$ 
in G\"odel's theorem is not provable in the formal
system.} 
\equ
 As described above, the 
 observing formula 
$\overline{\mbox{\App}}$ 
 can be interpreted as a formalization of
the proposition that \mbox{\App} is true from 
a mathematician's point of view
 who
 proves G\"odel's theorem, that is, 
 the proposition that \mbox{\App} is true
 can be formalized and be proved 
 in the observing formal system $\overline{\mbox{$S$}}$. 
 The formal symbol $\boldsymbol{S}$ for $S$ in $\overline{\mbox{$S$}}$ can be regarded as a reference to $S$ ``as a whole". 

According to (\ref{pGtS0}) in the proof of Theorem (\ref{tprovable}) the formula
\beq
\mbox{\em Consys} \implies \mbox{\App}, \label{fpGt}
\eeq
which is a formalization of the first part of G\"odel's theorem,
can be proved in $S$. But in this formalization (\ref{fpGt}) the 
reference to $S$ is lost, in particular, the reference of 
\mbox{\em Consys} to $S$. It cannot be added to $S$
because this implies that \mbox{\App} can be proved in $S$
which contradicts G\"odel's theorem.

The observing system $\overline{\mbox{$S$}}$ is a formal copy of $S$
containing a formal symbol $\boldsymbol{S}$ for $S$ (see Definition \ref{observing}).
Because $\overline{\mbox{$S$}}$ is a formalization of G\"odel's theorem and proof
from a mathematician's point of view
and $\overline{\mbox{\App}}$ cannot be proved in $S$,
$\overline{\mbox{$S$}}$, which is a formalization of the first part of G\"odel's theorem 
and proof including a reference to $S$, cannot be represented 
in $S$. This implies that $\overline{\mbox{$S$}}$ is more ``powerful" than $S$. Therefore,
a mathematician can construct a more ``powerful" formal system $\overline{\mbox{$S$}}$
from any consistent formal system $S$ because G\"odel's proof
applies to any consistent system $S$, that is, ``to any formalism whatever".
This means that Theorem \ref{tprovable} can be regarded as a proof of
the general existence principle, where 
the Turing program $P_i$ in the existence principle represents the system $S$
and $\psi$ in the existence principle corresponds to
the construction of $\overline{\mbox{$S$}}$ from $P_i$, that is, $S$.
Thus, the application of $\psi$ to the G\"odel number $i$ of $P_i$
yields $\overline{\mbox{$S$}}$ which contains a formal symbol $\boldsymbol{S}$ for $S$ and a formal proof of 
the first part of G\"odel's theorem
from a mathematician's point of view who refers to $S$.

Theorem \ref{tprovable} implies that 
 the logical existence
principle cannot be formalized, that is, it
implies the existence of an informal physical process which is described 
by the physical existence principle.
In view of Theorem \ref{tprovable} this process
can include a reference to any formal system $S$ 
which is represented in a more ``powerful" formal system $\overline{\mbox{$S$}}$
by a formal symbol $\boldsymbol{S}$ for $S$.
Thus, Theorem \ref{tprovable} implies the existence of 
physical systems that are capable of constructing 
 a more ``powerful" formal system $\overline{\mbox{$S$}}$
 from any formal system $S$ in an informal physical process.
 These systems are called creative.
 
 The general existence principle is a 
 generalization of the existence principle in Section \ref{SExistence}
 which is restricted to productive functions $\psi$ for productive sets $A$
 that are applied to the G\"odel number $i$ of a Turing program $P_i$
 whose output is a subset of $A$.
 A creative system (see Definition \ref{creativesystem} in Section \ref{SExistence})
 can apply $\psi$ to the G\"odel number $i$ of any $P_i$ although
 this application cannot be formalized because $\psi$ 
 is a productive function. The reason is that a general formal
 reference to the G\"odel number $i$ cannot be represented within $P_i$.
 This corresponds to the impossibility 
 to represent a reference to $S$ within a formalization (\ref{fpGt})
 of the first part of G\"odel's theorem in $S$.
 This reference can only be represented in an extended formal system $\overline{\mbox{$S$}}$
 that contains a formal symbol $\boldsymbol{S}$ for $S$ representing this reference.

For decades there is a discussion whether 
G\"odel’s incompleteness theorem implies limitations
 on what computers can prove (see Lucas \shortcite{Lucas61}).
Russell and Norvig \shortcite[p.\ 1023]{Russell...10} write:
\bqu 
Philosophers such as J. R. Lucas (1961) have claimed that this theorem shows that machines
are mentally inferior to humans, because machines are formal systems that are limited by the
incompleteness theorem - they cannot establish the truth of their own G\"odel sentence - while
humans have no such limitation. 

... it is impossible to prove that humans are not subject to G\"odel’s incompleteness 
theorem because any rigorous proof would require a formalization
of the claimed unformalizable
human talent, and hence refute itself.
\equ
Theorem \ref{tprovable} states that the observing formula 
$\overline{\mbox{\App}}$
is provable in the observing formal system $\overline{\mbox{$S$}}$.
 As described above, 
$\overline{\mbox{\App}}$
 can be interpreted as a formalization of
the proposition that the undecidable formula \mbox{\App} in G\"odel's theorem 
is true from 
a mathematician's point of view
 who
 proves G\"odel's theorem.
The proof of Theorem \ref{tprovable} can be regarded as a 
rigorous proof that the construction of the formal system 
system $\overline{\mbox{$S$}}$ from any system $S$ cannot be formalized.
$\overline{\mbox{$S$}}$ is a formalization of the proof of the first part of G\"odel's 
(first incompleteness) theorem from a mathematicians point of view. This formalization
is used in the proof of G\"odel's second incompleteness theorem.
Furthermore, $\overline{\mbox{$S$}}$ contains a formal symbol $\boldsymbol{S}$ for $S$
which is a formalization of the reference to $S$ of a mathematician 
who proves the first part of G\"odel's (first incompleteness) theorem.
This reference cannot be formalized in $S$ because, as described above, 
a formalization of this reference yields a contradiction to
 G\"odel's (first incompleteness) theorem.

According to Theorem \ref{tprovable}
the observing formula  $\overline{\mbox{\App}}$ of 
undecidable formula \mbox{\App} in a formal system $S$
is provable 
in the observing system $\overline{\mbox{$S$}}$ of $S$ if $S$ is consistent.
We write $S_0$ for $S$ and $S_1$ for $\overline{\mbox{$S$}}$.
$S_1$, that is, $\overline{\mbox{$S$}}$, contains another undecidable formula whose observing formula
can be proved in another observing system $S_2$, and so on.
Thus, a sequence $S_1$, %
$S_2$, $S_3$, ... of more and more ``powerful"
formal systems arises which contain proofs of more and more %
formulas. 
As described above Theorem \ref{tprovable} implies that
each of the formal systems
$S_i$ cannot be represented 
in the preceding formal system $S_{i-1}$. %
By means of G\"odel numbers the proofs of more and more %
formulas in $S_1$, $S_2$, $S_3$, ... can be transformed into
proofs in number theory. Thus, a more and more complete number theory arises.

Church's \shortcite[pp.\ 90, 100-102]{Church35} thesis\footnote{The term {\em Church's thesis} is due to
Kleene \shortcite[p.\ 274]{Kleene43} (see Kleene \shortcite[pp.\ 300, 317]{Kleene52}).}
 states that every effectively calculable function
is general recursive, that is, computable by a Turing machine (see 
Church, \citeyear[pp.\ 90, 100-102]{Church35}, and
Kleene \shortcite[pp.\ 300--301, 317--323]{Kleene52}).
Since ``effective calculability" is an intuitive concept,
the thesis cannot be proved (see Kleene 
\shortcite[p.\ 317]{Kleene52}).\footnote{In his article 
``Why G\"odel Didn't Have Church's Thesis"
Davis \shortcite[p.\ 22, footnote 26]{Davis82}
writes: ``We are not concerned here with attempts to distinguish
'mechanical procedures' (to which Church's thesis is held to apply)
from a possible broader class of 'effective procedures' ..."}

Church \shortcite[pp.\ 90, 102]{Church35} presents his thesis as a ``definition
of effective calculability:
\bqu
... (1) by defining a function to be effectively calculable
if there is an algorithm for the calculation of its values ...
\equ
According to Theorem \ref{tphysical} in Section \ref{SExistence}
the sequence (\ref{tphysicalseq})
produced by
a creative system $C$ that applies the physical rule 
(\ref{ruleexists}) cannot be computed by any Turing program.

Church \shortcite[pp.\ 90, 102]{Church35} 
proposes a second definition
of effective calculability:
\bqu
... (2) by defining a function $F$ (of positive integers) to be effectively calculable
if, for every positive integer $m$,
there exists a positive integer $n$
such that $F(m)=n$ is a provable theorem.
\equ
If we require for every Turing program $P_i$ in the input of the physical rule  (\ref{ruleexists})
in Section \ref{SExistence}
a proof in a formal system that the Turing program $P_i$ computes a total
function whose output is a subset of a productive set $A$,
then, for every G\"odel number (positive integer) $i$
in the input of  (\ref{ruleexists}), whose output is the sequence (\ref{tphysicalseq})
in Theorem \ref{tphysical},
there exists a natural number $y$
such that $\psi(i)=y$ is a provable theorem in some 
formal system, say $S_i$.
Such a formal system also exists 
for any finite set of natural numbers $i$
in the input of (\ref{ruleexists}).
But, because of Theorem \ref{tphysical}, there exists no formal system $S$,
which can be represented by a Turing program,
such that $\psi(i)=y$, where $y$ is a natural number, is a provable theorem in $S$
for all G\"odel numbers $i$ of Turing programs $P_i$ in the input of (\ref{ruleexists}).
Roughly speaking,  Theorem \ref{tphysical} implies that
the formal systems $S_i$ cannot be unified into a single formal system $S$.

In a
letter of June 8, 1937, to Pepis
Church wrote (see
Sieg \shortcite[pp.\ 175--176]{Sieg97}): 
\bqu
... if a
numerical function $f$ is effectively calculable then for every positive
integer $a$ there must exist a positive integer $b$ such that a valid proof
can be given of the proposition $f(a) = b$ ...

Therefore to discover a function which was effectively calculable
but not general recursive would imply discovery of an utterly new
principle of logic, not only never before formulated, but never before
actually used in a mathematical proof - since all extant mathematics is
formalizable within the system of Principia [Mathematica], or at least within one of
its known extensions. Moreover this new principle of logic must be of
so strange, and presumably complicated, a kind that its metamathematical
expression as a rule of inference was not general recursive
 ..." .
 \equ
 The proof of Theorem \ref{tphysical} in Section \ref{SExistence} uses
 the physical rule (\ref{ruleexists}) which is an implementation of the existence 
 principle.
As far as we know
the existence principle was 
``never before
actually used in a mathematical proof".
It allows the application of productive functions
to any existing Turing program.  
This cannot be achieved by formal systems which cannot apply
productive functions to the recursively enumerable subsets of productive
sets with which they deal only incompletely.
The logical existence principle abstracts the physical processes
from the physical existence principle such that it
can be regarded as a principle of logic 
which provides a general method of self-reference and self-application
that cannot be formalized.
The output (\ref{tphysicalseq}) of the physical rule  (\ref{ruleexists})
in Theorem \ref{tphysical}, which is an implementation of the existence principle, is
not general recursive, that is, it cannot be computed by a Turing program.
The existence principle may be regarded as strange because it refers to the perception of the physical existence of a Turing program as a whole.

This work was influenced by Post \shortcite{Post65}.
For example, Post \shortcite[p.~417]{Post65} writes:
\bqu
\bc
The Logical Process is Essentially Creative
\ec
This conclusion, so in line with Bergson's ``Creative Evolution",
 ... We see that a machine would never give a complete logic; for
once the machine is made we could prove a theorem it does not prove.
\equ
In %
``Creative Evolution" Bergson \shortcite[p.~342]{Bergson11} writes:
\bqu\it
Time is invention or it is nothing at all.
\equ
The formal system $S$ in Theorem \ref{tprovable} can be regarded as a ``machine" that
cannot prove the undecidable formula \App.
Theorem \ref{tprovable} states that the observing formula $\overline{\mbox{\App}}$
of \App\ can be proved in the observing formal system $\overline{\mbox{$S$}}$ which is a copy
of $S$ containing a formal symbol $\boldsymbol{S}$ for $S$.
As described above, the observing formula $\overline{\mbox{\App}}$
can be regarded as a formalization of the proposition that the undecidable formula \App\
is ``true".
A mathematician is capable of referring to the ``machine" $S$ as soon as $S$
``is made".  %
Theorem \ref{tprovable} implies that this
is achieved in a physical process that cannot be formalized in a 
single formal system.

Theorem \ref{tprovable} confirms Maturana and Varela \shortcite{MaturanaVarela80}. 
For example,
Maturana and Varela \shortcite[ p. 51]{MaturanaVarela80} write:
\bqu
... he [the observer] both creates (invents) relations and generates (specifies) the world (domain of interactions) in which he lives by continuously expanding his cognitive domain through recursive {\em descriptions} and representations of his interactions. 
The {\em new}, then, is a necessary result of the historical organization of the observer that makes of every attained state the starting point for the specification of the next one, which thus cannot be a strict repetition of any previous state; creativity is the cultural expression of this unavoidable feature. 
\equ
As described above, a sequence $S_1$, $S_2$, $S_3$, ... of more and more ``powerful"
observing systems 
can be produced in an informal process according to 
the general existence principle.
The observing systems $S_1$, %
$S_2$, $S_3$, ... can be regarded as recursive descriptions
of an observer because each $S_i$ is constructed from the preceding system $S_{i-1}$
and a formal symbol for $S_{i-1}$.
Thus, every attained state, that is, $S_{i-1}$, is the starting point for the specification 
of the next one, that is, $S_i$, which cannot be a strict repetition of any previous state,
for example, $S_{i-1}$,
because Theorem \ref{tprovable} implies that
each of the formal systems
$S_i$ cannot represented 
in the preceding formal system $S_{i-1}$.
Maturana and Varela \shortcite[ p. 53]{MaturanaVarela80}
write: %
\bqu
We cannot speak about the substratum in which our cognitive behavior is given, and about that of which we cannot speak, we must remain silent, as indicated by Wittgenstein. ... It means that we recognize that we, as thinking systems, live in a domain of descriptions, ..., and that through descriptions we can indefinitely increase the complexity of our cognitive domain. 
\equ
As described above, Theorem \ref{tprovable} implies that the construction of 
the observing system $\overline{\mbox{$S$}}$ from the formal system $S$ 
cannot be represented in $S$,
that is, there is no general formal description of the construction
of the formal system $\overline{\mbox{$S$}}$ from $S$.\footnote{Ammon \shortcite[Section 3.4]{Ammon87} introduces
principles on creative processes that are based on computer experiments. For 
example, the shunyata principle in  Ammon \shortcite[Section 3.4.5]{Ammon87} states that there is no explicit and general description of
creative processes.
}
Maturana and Varela \shortcite[p. 242]{MaturanaVarela87} write: 
\bqu
By existing, we generate cognitive ``blind spots"
that can be cleared only through generating new blind spots
in another domain. 
We do not see what we do not see, and what we do not see does not exist.
\equ
The undecidable formula \App\ in G\"odel's theorem
can be regarded as a ``blind spot" in 
a formal system, say $S$. 
According to  Theorem \ref{tprovable}, this ``blind spot" is ``cleared"
by the observing formula 
$\overline{\mbox{\App}}$ in
the observing system $\overline{\mbox{$S$}}$, which is a copy of $S$
containing a formal symbol $\boldsymbol{S}$ for $S$. 
This means that the observing formula 
$\overline{\mbox{\App}}$, which is a formalization 
of the proposition that \App\ is ``true",
can be proved in the 
observing system $\overline{\mbox{$S$}}$, that is, in another ``domain".
As described above, 
a reference to $S$ cannot be represented in  $S$, that is,
the existence of the formal system $S$ cannot be represented 
in $S$. In this sense, $S$ cannot represent its own existence.
This reference to $S$ is represented in $\overline{\mbox{$S$}}$
by a formal symbol $\boldsymbol{S}$ for $S$ (see Definition \ref{observing}).
As described above, 
Theorem \ref{tprovable} implies that this reference to $S$ as a whole
is achieved in a physical process that cannot be formalized in a 
single formal system.
Maturana and Varela \shortcite[p. 224]{MaturanaVarela87}
describe an experiment with a gorilla that suggests the existence of 
a physical reference of the gorilla to himself as a whole:
\bqu
A gorilla ... when first confronted with a mirror
will appear amazed and interested,
but after becoming used to it, he will ignore it.
... experimenters anesthetized a gorilla.
A colored dot was painted between his eyes - a place that could be seen 
only in the mirror.
After awakening from anesthesia, he was given a mirror. What a surprise!
Her put his hand to his forehead to touch the colored dot.
... this experiment suggests that the gorilla can generate a domain of self
through social distinctions. In this domain there is a possibility of 
reflection as with a mirror or with language.
\equ
This experiment suggests that the gorilla has a fixed ``symbol" for himself,  that is, 
any fixed means of representation in his brain that refers to himself.
This reference to himself ``as a whole" is generated physically when he sees himself in the mirror. 
It causes him to put his hand to his forehead.
Thus, its own existence generates a physical reference between a
``symbol" for himself  in his brain and himself ``as a whole" and a
physical causation of an action. 
This confirms Maturana and Varela's view because
it suggests that the gorilla has a preliminary stage
of a possibility of reflection in the sense
that he can refer to himself ``as a whole".  
Theorem \ref{tprovable} implies that such as physical reference 
cannot be formalized in a single formal system. 
The construction of the observing system $\overline{\mbox{$S$}}$ from the formal system $S$ can 
be regarded as a model of reflection 
because $\overline{\mbox{$S$}}$ is a copy of $S$ that contains
a formal symbol 
 $\boldsymbol{S}$ referring to $S$ as a whole.

Theorem \ref{tprovable} states 
the observed formula of the undecidable formula in a formal system $S$ 
can be proved in another formal system $\overline{\mbox{$S$}}$ which 
is a copy of $S$ and contains
a formal symbol 
 $\boldsymbol{S}$ referring to $S$ as a whole.
The claim that G\"odel's theorem applies to any formal system
 implies that G\"odel's theorem also applies to $\overline{\mbox{$S$}}$, that is, it is possible
 to refer to $\overline{\mbox{$S$}}$ and apply G\"odel's proof to
 $\overline{\mbox{$S$}}$ although there is no general formal reference 
 to $\overline{\mbox{$S$}}$, in particular, no general formal procedure for the application 
 of  G\"odel's proof to
 $\overline{\mbox{$S$}}$, that is, to any observing formal system. The possibility of a reference
 to $\overline{\mbox{$S$}}$ as a whole, in particular, an application 
 of  G\"odel's proof to
 $\overline{\mbox{$S$}}$, that is, to any %
 formal system, is described in the general existence principle.
 
Ammon \shortcite[Sections 4 and 5]{Ammon93}
 describes a computer proof of G\"odel's theorem for any formal number theory $T$,
 that is, $T$ corresponds to the formal system $S$ in this paper.
 This computer proof of G\"odel's theorem 
 can be represented in another formal system, say $U$.
The claim that G\"odel's theorem applies to any formal number theory
implies that G\"odel's theorem also applies to $U$ although  
there is no general formal procedure for the application 
 of  G\"odel's proof to $U$, that is, any formal system in which G\"odel's
 proof is represented.
As described above, a reason is that
 such a system cannot refer to itself 
as a whole.
  Referring 
  to Penrose \shortcite[p.\ 694]{Penrose90} Russell and Norvig \shortcite[p.\ 826]{Russell...95}
write:
\bqu
Penrose does not say why he thinks the "G\"odelian insight" 
[that the ``G\"odel sentence" $G(F)$ of a formal system $F$ is true] is not formalizable, and it appears
that in fact it has been formalized. In his Ph.D. thesis, Natarajan Shankar (1986) used the
Boyer-Moore theorem prover BMTP to derive G\"odel's theorem from a set of basic axioms, in
much the same way that G\"odel himself did.$^8$ 

...

$^8$ Ammon's SHUNYATA system (1993) even appears to have developed by itself the diagonalization technique used by
G\"odel and developed originally by Cantor.
\equ
Because there is no general formal procedure for the application 
 of  G\"odel's proof to any formal system in which G\"odel's
 proof is represented, a formal proof of G\"odel's theorem
 cannot be general.\footnote{Ammon's SHUNYATA program \shortcite{Ammon93} generated a proof of G\"odel's 
theorem in the form: 
There is a closed formula $F$ in any formal number theory $T$
such that
\bi
\ii if the theory $T$ is consistent, 
$F$ is not provable in $T$, and
\ii if the theory $T$ is $\omega$-consistent, 
$\neg F$ is not provable in $T$
\ei
whose formalization in Ammon \shortcite{Ammon93} is:
\bc
\bt {l} 
$\all \; T (\fnt (T) \rightarrow \ex \, F  (\cf (F,T)$ \& \\ [0cm]
\hspace*{3.5cm} $(\consistent (T) \rightarrow \nt (\provable (F, T)))$ \&
								\\ [0cm]
\hspace*{3.5cm} $(\omegaconsistent (T) \rightarrow
				\nt (\provable (\neg F, T)))))$. \\ [0cm]
\et 
\ec
 In the formalization $\fnt (T)$ means that $T$
 is a formal number theory. Thus, the formalization of G\"odel's theorem
 in  Ammon \shortcite{Ammon93} refers to the [incomplete] formal theory $T$.
 Because the formal theory, say $U$, in which G\"odel's theorem and proof 
 in  Ammon \shortcite{Ammon93} are
 represented, is incomplete as well, the claim that G\"odel's
 theorem applies to all formal number theories $T$ implies that
 the theorem also applies to $U$.
 But the applicability  of G\"odel's
 theorem to $U$  cannot be represented in $U$ because $U$ cannot refer to $U$ itself,
 that is, to itself as a whole.
 This implies that there is no general formalization 
 of the reference to all formal theories $T$ to which G\"odel's theorem 
 applies.
 }

 As described above, 
the possibility to apply any Turing-computable
function to any Turing program satisfying some properties, in particular,
 G\"odel's proof to
 any formal system 
 in which G\"odel's
 proof is represented,
  is characterized in the general existence principle.

G\"odel's %
theorem states that any formal system, say $S$, satisfying some simple properties contains an undecidable formula \App, that is, neither \App\ nor its negation  $\neg \mbox{\App}$ can be proved in %
$S$ (see Kleene \shortcite[p.\ 207, Theorem 28]{Kleene52}).
The ``G\"odelian insight" in
Russell and Norvig \shortcite[p.\ 826]{Russell...95} corresponds to 
Kleene \shortcite[p.\ 426]{Kleene52}): ``... we can recognize that ${\rm A}_p(\boldsymbol{p})$ 
is true by taking into view the structure of that [formal] system as a whole".
\App, that is, 
$\forall {\rm \textsl{b}} \, \neg {\rm A}(\boldsymbol{p}, \textsl{b})$, is %
a formalization of the
proposition 
that the formula 
\App \ is not provable in the formal
system $S$
because 
$b$ is the G\"odel number of a proof of the formula ${\rm A}_a(\boldsymbol{a})$ %
in the definition of the predicate $A(a,b)$ (see Kleene \shortcite[p.\ 206, Lemma 21]{Kleene52})
which is expressed by the formula ${\rm A}(\textsl{a}, \textsl{b})$ 
(see Kleene \shortcite[p.\ 207]{Kleene52}).

Thus, \App\ is  
a formalization of the ``true"
proposition 
that %
\App\ is not provable in $S$ 
although 
\App\ 
is not provable in $S$ to according G\"odel's theorem. 
Roughly speaking,
\App\ is ``true" but not provable in $S$.

The undecidable formula \App\ in the incomplete formal system $S$ 
does not contain a %
reference to $S$.
If we add a formal symbol $\boldsymbol{S}$ for $S$ to 
\App\
we obtain the observing formula  $\overline{\mbox{\App}}$
of \App\
(see Definition \ref{observing} and Example \ref{Apstrokepstroke}).
$\overline{\mbox{\App}}$
is a formalization of the proposition that
\App\ is not provable in $S$.
Theorem \ref{tprovable} states that the observing formula 
 $\overline{\mbox{\App}}$ is provable in the observing [formal] system~$\overline{\mbox{$S$}}$.
All observing formulas of formulas that are provable in $S$
are provable in the observing system $\overline{\mbox{$S$}}$. Additionally, the observing formula
$\overline{\mbox{\App}}$ of \App\ is provable in $\overline{\mbox{$S$}}$.
Therefore, the observing  system $\overline{\mbox{$S$}}$ is more powerful in the sense that it can prove more theorems than the observed system $S$.

The existence principle in Section \ref{SExistence},
which is restricted to productive functions,
  is 
a special case of the 
general existence principle in Section \ref{SDiscussion}
which applies to any Turing-computable function.

Roughly speaking, the general existence principle is the core of Turing's residue which is called
initiative by Turing (see Section 1).

\section{Conclusion}
The physical existence principle states that the
perception of the physical existence of any Turing program
can serve as a physical causation for the application of Turing-computable functions
to this Turing program.
The logical existence principle abstracts the logical aspects
from the physical existence principle.
It is used in the proof of a theorem stating that
a physical rule, which is an implementation of the existence
principle, produces a sequence that cannot be computed by any Turing program.
The logical existence principle can be regarded as a new principle of logic that
was never before
used in a mathematical proof.
The existence principle overcomes the incompleteness of formal systems
and the limits of Turing machines because it describes 
the perception of the physical existence of Turing programs as a whole.
This allows a general reference and application of formal systems to themselves as a whole which cannot be achieved
within formal systems themselves.
The generality of G\"odel's theorem implies its applicability 
to formalizations of its proof. There is no general formal procedure
for this applicability because formalizations of G\"odel's proof
are incomplete as well, that is, these formalizations are not general because
they cannot include the applicability of G\"odel's theorem and proof
to themselves. The general existence principle describes this informal aspect 
of G\"odel's theorem and proof.
A physical system that contains an implementation of the physical existence principle is 
called creative system.
Creative systems can prove the observing formula of G\"odel's undecidable
formula in an observing formal system which contains a copy of the incomplete formal
system and a formal symbol referring to the incomplete formal system.
Therefore, the observing  system is more powerful in the sense that it can prove
more theorems than the observed system.

\bigskip \medskip %
\noindent
{\bf Acknowledgments}. The author wishes to thank Andreas Keller
 for helpful comments on earlier versions of 
this paper and many people for their
interest, encouragement, and support in the course of many years.

\bibliography{Physical}
\bibliographystyle{named} %

\end{document}